\documentclass[A4,conference]{IEEEtran}

\usepackage{calc}
\usepackage{stfloats}
\usepackage{calc}
\usepackage{microtype}
\usepackage[intlimits,sumlimits,namelimits]{amsmath} 
\everymath{\displaystyle} 
\usepackage{amssymb}
\usepackage{amsfonts}
\usepackage{psfrag}
\usepackage{verbatim}
\usepackage{multirow}
\usepackage{multicol}
\usepackage{algorithm2e}
\usepackage{blkarray}
\usepackage{wrapfig}
\usepackage{paralist}
\usepackage{trfsigns}
\usepackage{version}
\usepackage{wasysym}
\usepackage{bm}
\usepackage{color}
\usepackage{xcolor}
\usepackage{graphicx}
\usepackage[most]{tcolorbox}
\usepackage[export]{adjustbox} 
\usepackage{caption}
\usepackage{svg}
\svgsetup{inkscapelatex=false,inkscapepath=Graphics/generatedPDF}
\usepackage[
    backend=biber,
    style=numeric,
    doi=false,
    isbn=false,
    url=false,
    eprint=false
]{biblatex}
\renewbibmacro{in:}{}
\appto\bibsetup{\flushbottom}
\AtBeginBibliography{\small}
\usepackage{flushend}

\def\vec#1{\underline{#1}}

\def\1_2{{\frac{1}{2}}}

\def\rv#1{\boldsymbol{#1}}

\DeclareMathOperator{\xlog}{xlog}

\DeclareMathOperator{\N}{N}
\DeclareMathOperator*{\argmin}{arg\,min}

\def\NewR{\mathbb{R}} 

\def\Eq#1{(\ref{#1})}

\def\Sec#1{Sec.~\ref{#1}}
\def\SubSec#1{Subsec.~\ref{#1}}

\def\Fig#1{Fig.~\ref{#1}}

\def\Remark#1{Remark~\ref{#1}}

\newlength\EqLen

\def\ScaleInner#1{%
  \settowidth{\EqLen}{#1}
  \ifdim\EqLen < \columnwidth%
    \begin{equation*}%
      \begin{minipage}{\EqLen}#1\end{minipage}%
    \end{equation*}%
  \else%
    \begin{equation*}%
      \resizebox{0.99\columnwidth}{!}{\begin{minipage}{\EqLen}#1\end{minipage}}%
    \end{equation*}%
  \fi%
}%

\def\Scale#1
{
  \ScaleInner{$ #1 $}
}

\def\LongVersion#1{}

\def\citep#1{(\cite{#1})}

\usepackage{amsthm}



\newtheorem{remarkx}{Remark}[section]
\newenvironment{remark}
{\remarkx}
{\endremarkx}

\makeatletter
\newcommand\SaveEquation[2]{\@namedef{equation@#1}{#2}}
\newcommand\UseEquation[1]{\@nameuse{equation@#1}}
\makeatother

\SetKw{KwBy}{by}

\SetCommentSty{mycommfont}

\usepackage{enotez}
\DeclareInstance{enotez-list}{sverre}{paragraph}{
  heading=\section*{#1},
  notes-sep=\baselineskip,
  format=\normalsize\normalfont\leftskip1.8em,
  number=\makebox[0pt][r]{#1.\quad}\ignorespaces,
}

\newtcolorbox{YellowBox}{
  enhanced,
  boxrule=0pt,frame hidden,
  borderline east={1mm}{0pt}{yellow!90!black},
  borderline west={1mm}{0pt}{yellow!90!black},
  colback=yellow!40!white,
  sharp corners,
  grow sidewards by=1.5mm,
  top=0mm,
  bottom=0mm,
  left*=0mm,
  right*=0mm
}

\date{}

\makeatletter
\def\section{\@startsection{section}{1}{\z@}{-1.5ex plus -1ex minus -0.5ex}%
    {0.7ex plus 0.5ex minus 0ex}{\normalfont\normalsize\centering\scshape}}%
\def\subsection{\@startsection{subsection}{2}{\z@}{-1ex plus -0.5ex minus -0.5ex}%
    {0.5ex plus 0.1ex minus 0ex}{\normalfont\normalsize\itshape}}%
\makeatother

\author{\IEEEauthorblockN{\textbf{Uwe D.\ Hanebeck}}\\
\IEEEauthorblockA{Intelligent Sensor-Actuator-Systems Laboratory (ISAS)\\
Institute for Anthropomatics and Robotics\\
Karlsruhe Institute of Technology (KIT), Germany\\
Uwe.Hanebeck@kit.edu}}

\begin{document}

\title{Progressive Bayesian Particle Flows\\
based on Optimal Transport Map Sequences}

\maketitle
\thispagestyle{empty}
\pagestyle{empty}


\begin{abstract}
    %
%
We propose a method for optimal Bayesian filtering with deterministic particles.
%
%
In order to avoid particle degeneration, the filter step is not performed at once.
Instead, the particles progressively flow from prior to posterior.
This is achieved by splitting the filter step into a series of sub-steps.
%
%
In each sub-step, optimal resampling is done by a map that replaces non-equally weighted particles with equally weighted ones.
%
%
Inversions of the maps or monotonicity constraints are not required, greatly simplifying the procedure.
%
%
The parameters of the mapping network are optimized w.r.t.\ to a particle set distance.
This distance is differentiable, and compares non-equally and equally weighted particles.
%
%
Composition of the map sequence provides a final mapping from prior to posterior particles.
%
%
Radial basis function neural networks are used as maps.
%
%
It is important that no intermediate continuous density representation is required.
The entire flow works directly with particle representations.
This avoids costly density estimation.
\end{abstract}


\section{Introduction} \label{Sec_Intro}

%
%
We consider Bayesian filtering based on particle representations of probability density functions (pdfs).
%
%
Particle filters are a very active field of research.
Thousands of papers have been devoted to this topic.
%
%
Most of them focus on large numbers of random particles and asymptotically valid methods.
Necessarily finite numbers of particles lead to a variety of problems.
This includes (i)~particle degeneracy and (ii)~non-reproducibility due to the randomness of the particles.

%
%
Standard approaches for coping with degeneracy employ \emph{resampling}.
By doing so, the weighted samples are replaced by a set of unweighted samples in order to reduce sample variance.
%
%
However, standard resampling procedures simply delete small samples and replicate large ones. 
Their locations remain unchanged until the next prediction step. 
This is clearly suboptimal and does not allow for several consecutive filtering steps without intermediate prediction steps.
%
%
In addition, Proposal densities and importance sampling are used to alleviate degeneration effects.
This allows to consider the current measurement during prediction and places the particles closer to the likelihood.
However, this requires additional engineering and increases computational complexity.

%
%
We focus on methods that employ different forms of Bayesian filter updates.
They replace the direct update by an indirect one that inherently avoids degeneration by using transport maps or flows.

\section{State of the Art} \label{Sec_State_of_Art}

%
%
The state of the art in calculating transport maps or flows will be investigated.
The focus is on methods for converting the Bayesian filter step.
However, some methods are only derived for the case of mapping between two given densities without explicitly considering the Bayes update.

\subsection{Continuous Density Flow Filters}

%
%
We start with flows of continuous densities as they have been developed earlier than particle flows.
%
%
The first progressive Bayesian filter \cite{hanebeckProgressiveBayesNew2003} employs a continuous density for representing the posterior.
Progressively introducing the likelihood function leads to a homotopy continuation approach.
A Gaussian mixture flow is derived that minimizes the squared integral deviation from the true posterior.
This results in a system of explicit ordinary first-order differential equations solved over an artificial time from $0$ to $1$.
The approach is generalized in \cite{hagmarOptimalParameterizationPosterior2011} for the Kullback-Leibler divergence and the squared Hellinger distance.


\subsection{Filters based on Transform Maps}

%
%
Transformation of a random vector $\rv{\vec{x}}$ via a known nonlinear map $\rv{\vec{y}} = \vec{g}(\rv{\vec{x}})$ is a classic problem.
The goal is to calculate the output density $f_y(\vec{y})$ given the input density $f_x(\vec{x})$.
For that purpose, a different map is required that maps densities to densities, i.e., $f_y = G(f_x)$.
It can be derived from the original mapping $\vec{g}(\cdot)$, its roots, and its Jacobian.
The derivation is simplified for monotonic maps.

%
%
Here, we are faced with a more complex problem.
Given two densities $f_x(\vec{y})$ and $f_y(\vec{y})$, we \emph{want to find the map} $\vec{g}(.)$ between their corresponding random vectors $\rv{\vec{x}}$ and $\rv{\vec{y}}$.
When the two densities are continuous, a mapping exists, is unique, and monotonic \cite{mccannExistenceUniquenessMonotone1995}.
Finding the map is challenging, especially under the monotonicity constraint.
In univariate settings, the map can be composed from the input cumulative distribution and the output quantile function.
However, calculation of cumulative distributions and quantile functions can be challenging with analytic expression only available in special cases.
This is exacerbated in multivariate settings.

%
%
It gets even more challenging, when only samples of the two densities $f_x(\vec{y})$ and $f_y(\vec{y})$ are given.
In that case, standard distance measures (such as the KL divergence) between densities cannot be employed.
The most complicated case is the Bayes update, when only samples are given for $f_x(\vec{y})$.
Then $f_y(\vec{y})$ is only given implicitly as the product of the samples and the likelihood and we can neither obtain values of the posterior nor can we sample from it.

%
%
\paragraph*{Filters based on Knothe-Rosenblatt Triangular Maps}%
In \cite{elmoselhyBayesianInferenceOptimal2012}, a continuous prior density is assumed to be given from which samples can easily be drawn.
In addition, a given likelihood is assumed that can be evaluated up to a constant.
A map is characterized by the minimum of either the Hellinger metric or the KL divergence between prior and posterior, see \cite[7818-7819]{elmoselhyBayesianInferenceOptimal2012}.
The map is calculated by minimization, which is a non-convex problem, where the distance is evaluated by Monte Carlo simulation from then prior.
Knothe-Rosenblatt rearrangements \cite{knotheContributionsTheoryConvex1957} are used as these are triangular and monotonic and thus easy to invert\footnote{For general triangular transformations, see \cite{bogachevTriangularTransformationsMeasures2005}.}.
It is important to note that these maps are not optimal with respect to a transportation distance.
To reduce complexity, decomposability of transport maps is investigated in \cite{spantiniInferenceLowDimensionalCouplings2018a}, leading to sparse triangular maps.
%
%
The construction of maps is generalized in \cite{marzoukSamplingMeasureTransport2016}, where two types of maps are considered.
The first map\footnote{This is the map used in \cite{elmoselhyBayesianInferenceOptimal2012}.} (direct transport) transforms a reference measure to a target measure
\cite[6]{marzoukSamplingMeasureTransport2016}.
The reference density is known, the target density can be evaluated but is unnormalized.
The second map (inverse transport) transforms the target measure to the given reference measure \cite[13]{marzoukSamplingMeasureTransport2016}.
This is useful when the target density is unknown and only samples of it are given.
Finding the map is equivalent to maximum likelihood estimation and can be solved via convex optimization.
%
%
These two maps are used in \cite{spantiniCouplingTechniquesNonlinear2019} to derive a nonlinear ensemble filter \cite[20]{spantiniCouplingTechniquesNonlinear2019} for the case when only samples of the prior density are available:
An inverse map is constructed for transforming prior samples to a convenient reference measure.
A direct map is used to transform the reference measure to the posterior.

%
%
\paragraph*{Filters based on Normalizing Flows}
Popularized in the machine learning community, normalizing flows are used in variational inference problems \cite{rezendeVariationalInferenceNormalizing2015} to perform density estimation in order to model complex data distributions.
A normalizing flow is a sequence of invertible transformations mapping a reference measure to a set of samples from a desired target density
\cite[3]{papamakariosNormalizingFlowsProbabilistic2021}.
The flows correspond to the inverse transport in \cite{marzoukSamplingMeasureTransport2016} and are determined by maximum likelihood estimation.

\subsection{Particle Flow Filters: Continuous Derivation}

%
%
The first breed of particle flow filters derive a PDE or ODE based on a suitably parametrized posterior while assuming continuity of the involved densities.
In a second step, the required discretizations are performed.

\paragraph*{Daum-Huang (DH) Particle Flows}
Particle flows are derived in \cite{daumParticleFlowNonlinear2008} from a log-homotopy relating prior and posterior.
The flow is represented by a partial differential equation (Fokker-Planck) assuming continuous densities.
Hence, estimating the required gradients from the particles is a challenging problem, see \cite{daumGradientEstimationParticle2009}.
Several versions of DH flows have been proposed \cite{daumSevenDubiousMethods2014}, many for coping with stiffness in the flow.
However, they usually rely on some sort of nonlinear Kalman filter running in parallel that compromises performance \cite{dingImplementationDaumHuangExactFlow2012}.

\paragraph*{Particle Flows based on Liouville Equation}
In \cite{hengGibbsFlowApproximate2021}, a homotopy continuation approach similar to \cite{hanebeckProgressiveBayesNew2003} is used.
Assuming an ODE for moving particles from prior to posterior, the corresponding Liouville PDE is derived%
\footnote{The Liouville PDE or continuity equation, is a special case  of the Fokker-Planck PDE for zero diffusivity.}.
(A similar approach is pursued in \cite[5]{meloStochasticParticleFlow2015}.)
The desired ODE velocity field is then obtained as the solution of the Liouville PDE.
Tractable solutions are obtained in the univariate case.
The multivariate case requires a Gibbs approximation.
This requires the full conditional distributions, which are numerically approximated.
Finally, the mapped particles are ``just'' used as proposal distributions inside of sequential Monte Carlo samplers.

\paragraph*{Particle Flows with Repulsion Kernels}
It would be possible to employ density estimation to find an intermediate continuous representation to calculate gradients.
However, density estimation usually is not differentiable.
In \cite{hanebeckFLUXProgressiveState2018}, repulsion kernels \cite{hanebeckKernelbasedDeterministicBluenoise2014} are used that
represent the spread of probability mass around the particles.
This leads to an ODE for the particle locations over an artificial time from $0$ to $1$.

\subsection{Particle Flow Filters: Direct Discrete Derivation}

%
%
The second breed of particle flow filters acknowledges that a computer implementation requires discretization anyway and directly derives a sequence of discrete updates.

%
%
In \cite{ruoffProgressiveCorrectionDeterministic2011}, the likelihood is adaptively split into several sub-likelihoods, each of which is easier to process.
The sub-likelihoods are used to sequentially update prior particles.
After each update, the weighted particles are optimally resampled with equally weighted ones.
Resampling is performed by minimizing a suitable particle set distance \cite{MFI08_Hanebeck-LCD} and \cite{hanebeckOptimalReductionMultivariate2015}.
As a result, the prior particles are moved to regions with high posterior density.
In \cite{hanebeckProgressiveBayesianEstimation2016}, smoothness assumptions are exploited to perform local up-sampling before resampling.

\section{Problem Formulation} \label{Sec_ProbForm}

%
%
We consider a dynamic system with a state $\vec{x} \in \NewR^D$ with state dimension $D$.
A transition density describing the state evolution provides a forecast.
This forecast is in the form of a prior density $f_p(\vec{x})$.

%
%
Given a prior density $f_p(\vec{x})$ and a likelihood function $f_L(\vec{x})$%
\footnote{The likelihood function is usually obtained by plugging a specific measurement, say $\hat{\vec{y}}$, into the conditional density $f(\vec{y} | \vec{x})$ describing the relation between measurement $\vec{y}$ and state $\vec{x}$ such that $f_L(\vec{x})=f(\hat{\vec{y}} | \vec{x})$.},
we consider a nonlinear Bayesian filter step for updating the prior density and calculating the posterior density $f_e(\vec{x})$ as
\begin{equation}
    f_e(\vec{x}) \propto f_p(\vec{x}) \cdot f_L(\vec{x}) \enspace .
    \label{Eq_BayesProp}
\end{equation}
%
%
The symbol $\propto$ indicates that a normalization is required as the multiplication of $f_p(.)$ and $f_L(.)$ does not automatically maintain normalized result.

%
%
We consider the important case of a prior density given purely as a set of samples (or particles) formally written as Dirac mixture density
\begin{equation}
    f_p(\vec{x}) = \sum_{i=1}^L w_{p,i} \cdot \delta(\vec{x}-\vec{x}_{p,i})
    \label{Eq_PriorDM}
\end{equation}
with weights $w_{p,i}>0$, $\textstyle\sum_{i=1}^L w_{p,i} = 1$, and sample locations $\vec{x}_{p,i}$.

%
%
For a given Dirac mixture prior, the Bayesian filter step becomes
\begin{equation}
    \begin{aligned}
        \tilde{f}_e(\vec{x}) & \propto f_L(\vec{x}) \cdot \sum_{i=1}^L w_{p,i} \cdot \delta(\vec{x}-\vec{x}_{p,i})                                     \\
                             & = \sum_{i=1}^L \underbrace{w_{p,i} \cdot f_L(\vec{x}_i)}_{\bar{w}_{e,i}} \cdot \delta(\vec{x}-\vec{x}_{p,i}) \enspace .
    \end{aligned}
\end{equation}
Upon normalization, we obtain the posterior weights $\tilde{w}_{e,i} = \bar{w}_{e,i} / \textstyle\sum_{i=1}^L \bar{w}_{e,i}$.
The posterior Dirac mixture is now given as
\begin{equation}
    \tilde{f}_e(\vec{x}) = \sum_{i=1}^L \tilde{w}_{e,i} \cdot \delta(\vec{x}-\tilde{\vec{x}}_{e,i}) \enspace .
    \label{Eq_PosteriorDM}
\end{equation}
The posterior sample locations do not change w.r.t.\ the prior samples, i.e., $\tilde{\vec{x}}_{e,i} = \vec{x}_{p,i}$ for $i=1,\ldots,L$.

%
%
The posterior $\tilde{f}_e(\vec{x})$ in \Eq{Eq_PosteriorDM} is derived from the straightforward application of the Bayesian filter step to Dirac mixtures.
This leads to a serious problem:
The samples are not equally weighted anymore and do not equally contribute to the representation of the posterior.
Often, some particle weights are (close to) zero, in fact dying out, leading to particle degeneracy mentioned above.
%
%
A typical scenario is large system noise, which spreads the particles during the prediction step combined with low measurement noise leading to narrow likelihoods.

%
%
Many solutions, some systematic, many of heuristic nature, have been proposed to solve the degeneracy problem,
%
%
which is a fundamental and difficult problem.

\begin{YellowBox}
    \begin{remark}
        Our goal is to derive a Bayesian filter that inherently avoids degeneracy without any heuristic approaches.
        It should be easy to understand, simple to implement, numerically stable, and robust.
    \end{remark}
\end{YellowBox}

%
%
Furthermore, we propose to use deterministic particles instead of random ones.
This (i)~reduces the required number of particles as the placement is more homogeneous and (2)~ensures reproducibility.
In this paper, we use the sampling method from \cite{hanebeckDiracMixtureApproximation2009} used in \cite{steinbringSmartSamplingKalman2016}.

\section{Optimal Resampling} \label{Sec_Resampling}

%
%
We now develop an optimal resampling step.
It replaces the non-equally weighted Dirac mixture $\tilde{f}_e(\vec{x})$ with its equally weighted approximation $f_e(\vec{x})$.
$\tilde{f}_e(\vec{x})$ is composed of non-equal weights $\tilde{w}_{e,i}$ and locations $\tilde{\vec{x}}_{e,i}=\vec{x}_{p,i}$.
$f_e(\vec{x})$ has equal weights, i.e., $w_{e,i} = w_{p,i} = 1/L$ and new locations $\vec{x}_{e,i}$.

%
%
We collect the weights and locations in sets for $f_e(\vec{x})$
\begin{equation}
    \begin{aligned}
        {\cal W}_e & = \left\{ w_{e,1}, w_{e,2}, \ldots, w_{e,L} \right\} \enspace ,                   \\
        {\cal X}_e & = \left\{ \vec{x}_{e,1}, \vec{x}_{e,2}, \ldots, \vec{x}_{e,L} \right\} \enspace ,
    \end{aligned}
\end{equation}
and for $\tilde{f}_e(\vec{x})$
\begin{equation}
    \begin{aligned}
        \tilde{{\cal W}}_e & = \left\{ \tilde{w}_{e,1}, \tilde{w}_{e,2}, \ldots, \tilde{w}_{e,L} \right\} \enspace ,                   \\
        \tilde{{\cal X}}_e & = \left\{ \tilde{\vec{x}}_{e,1}, \tilde{\vec{x}}_{e,2}, \ldots, \tilde{\vec{x}}_{e,L} \right\} \enspace ,
    \end{aligned}
\end{equation}
instead of vectors and matrices to underline that there is no inherent order.

%
%
\begin{YellowBox}
    \begin{remark}
        We first consider the case that no degeneration of  $\tilde{f}_e(\vec{x})$ occurred.
        This means that the weight variance in $\tilde{w}_{e,i}$ is small and all samples contribute to the density representation.
        Degeneration will be treated in \Sec{Sec_ProgressiveProcessing}.
        \label{Remark_NoDegeneration}
    \end{remark}
\end{YellowBox}

%
%
The key idea to finding $\vec{x}_{e,i}$ is to use an optimal map $\vec{M}(.)$ that transforms the prior random vector $\rv{\vec{x}}_p$ to the posterior random vector $\rv{\vec{x}}_e$
\begin{equation}
    \rv{\vec{x}}_e = \vec{M}(\rv{\vec{x}}_p) \enspace .
    \label{Eq_Map}
\end{equation}
This map is used to map prior samples $\vec{x}_{p,i}$ to posterior samples $\vec{x}_{e,i}$.
%
%
Mapping samples does not change their weights but their locations.
This guarantees equally weighted posterior samples $\vec{x}_{e,i}$.
%
%
We now have to find a map $\vec{M}(.)$ that leads to posterior samples that fulfill $f_e(\vec{x}) \approx \tilde{f}_e(\vec{x})$.

\begin{figure}[t]
    \begin{center}
        \includesvg{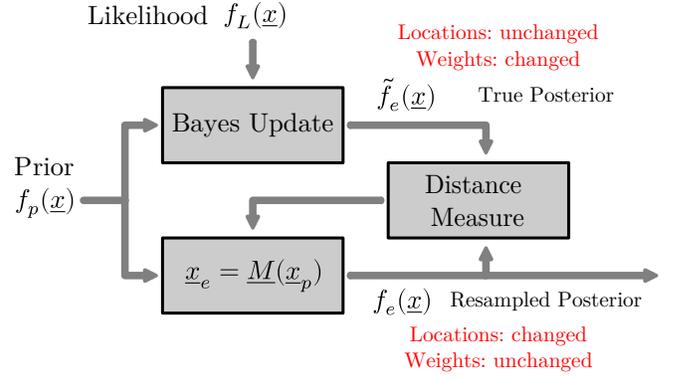}
    \end{center}
    \caption{%
        Block diagram of generating a resampling map given prior $f_p(\vec{x})$ and likelihood $f_L(\vec{x})$.
        $\tilde{f}_e(\vec{x})$ is the non-equally weighted posterior resulting from the base update.
        $f_e(\vec{x})$ is the resampled equally weighted posterior with $f_e(\vec{x}) \approx \tilde{f}_e(\vec{x})$.}
    \label{Fig_MapGeneration}
\end{figure}

%
%
The map generation is shown in \Fig{Fig_MapGeneration}.
$\tilde{f}_e(\vec{x})$ is obtained by a Bayes update, i.e., by multiplying $f_p(\vec{x})$ with the likelihood $f_L(\vec{x})$ (upper path).
It serves as the reference density.
Its locations $\tilde{\vec{x}}_{e,i}$ are identical to those of $f_p(\vec{x})$, only its weights $\tilde{w}_{e,i}$ are changed.
In the lower path, the map $\vec{M}(.)$ propagates $f_p(\vec{x})$ to $f_e(\vec{x})$.
$f_e(\vec{x})$ has identical weights as $f_p(\vec{x})$.
Its locations $\vec{x}_{e,i}$ have changed due to the mapping though.
We desire $f_e(\vec{x})$ to be close to $\tilde{f}_e(\vec{x})$ w.r.t.\ an appropriate distance measure $D$.
The map $\vec{M}(.)$ is adjusted accordingly by minimizing $D(f_e(\vec{x}),\tilde{f}_e(\vec{x}))$.

\subsection{Properties of Map} \label{SubSec_PropertiesMap}

%
%
It would be sufficient to use a discrete map for mapping the $L$ samples individually.
However, in that case the number of map parameters would be equal to the number of samples.
As a result, we would not have a complexity gain compared to a direct reapproximation.
In addition, there would be no smoothing.

%
%
Hence, we use a continuous map, preferably one with as few parameters as possible to reduce complexity.
%
%
This automatically allows interpolation between samples.
In summary, a few prior samples economically produce a smooth mapping that can be used to map many samples.

%
%
\begin{YellowBox}
    \begin{remark}
        Interpolation can be used to increase the number of posterior samples $\vec{x}_i^e$.
        This can be done after the map has been generated.
    \end{remark}
\end{YellowBox}
%
%
\noindent
The map $M(.)$ has the following properties:
\begin{itemize}
    \item[Property 1] Might be non-monotonic.
    \item[Property 2] No inverse required.
    \item[Property 3] Only used for mapping sample values.
    \item[Property 4] Differentiability w.r.t.\ parameters required.
    \item[Property 5] Differentiability not required w.r.t.\ $\vec{x}$.
\end{itemize}
%

\subsection{Specific Map}

%
%
According to \Remark{Remark_NoDegeneration}, in this section we assume that the change from $f_p(\vec{x})$ to $\tilde{f}_e(\vec{x})$ is small.
%
%
Hence, the map $\vec{M}(.)$ is close to the identity mapping.

%
%
Here, we propose a combination of an affine base map combined with a radial basis function nonlinearity.
For a single output it is given by
\begin{equation}
    M_i(\vec{x}) = \underbrace{\vec{a}_i^T \cdot \vec{x} + b_i}_{\text{affine part}}
    + \underbrace{\sum_{r=1}^R v_{r,i} \cdot \text{RBF}_r(\vec{x}-\vec{x}_r)}_{\text{radial basis function part}}
    \enspace ,
    \label{Eq_SpecificMap1D}
\end{equation}
with weights $v_{r,i} \in \NewR$, locations $\vec{x}_r$, and radial basis kernels $\text{RBF}_r(.)$.
%
%
The complete vector-valued map $\vec{M}(.)\colon \NewR^D \rightarrow \NewR^D$ is given by
\begin{equation}
    \vec{M}(\vec{x}) = \begin{bmatrix} M_1(\vec{x}), M_2(\vec{x}), \ldots, M_D(\vec{x}) \end{bmatrix}^T \enspace .
    \label{Eq_SpecificMap}
\end{equation}
%

%
%
For the optimization, initial values can simply be set to $\vec{a}_i=\vec{1}$, $b_i=0$, and $v_{r,i}=0$ for $r=1,\ldots,R$, $i=1,\ldots,D$.
The RBF locations $\vec{x}_r$ could be set to fixed a priori locations for $r=1,\ldots,R$.

\subsection{Distance Measure}

%
%
We have to compare the weighted Dirac mixture $\tilde{f}_e(\vec{x})$ with its unweighted counterpart $f_e(\vec{x})$.
A suitable distance measure should satisfy the following requirements:
%
%
\begin{itemize}
    \item[Requirement 1:] Handle Dirac mixture densities\footnote{i.e., discrete densities on a continuous domain.}.
    \item[Requirement 2:] Handle non-equal weights.
    \item[Requirement 3:] Handle non-equal supports.
    \item[Requirement 4:] Be differentiable w.r.t.\ locations (and weights).
\end{itemize}
%
%
The first three properties cannot be handled with standard distances that require continuous densities.
This includes the KL-divergence and integral squared distances.
%
%
On the other hand, Wasserstein distances could be used.
However, they suffer from large complexity.
In addition, property 4 is not fulfilled.
%
%
Here, we propose the use of the Cram{\'e}r-von Mises distance \cite{hanebeckOptimalReductionMultivariate2015} based on Localized Cumulative Distributions \cite{MFI08_Hanebeck-LCD}.

%
%
%
For two Dirac mixture densities $f_x$ with $L$ components, weights $w_{x,1}, w_{x,2}, \ldots, w_{x,L}$, locations
\begin{equation}
    \vec{x}_i = \begin{bmatrix} x_{1,i}, x_{2,i}, \ldots, x_{D,i} \end{bmatrix}^T \in \NewR^D
\end{equation}
for $i=1,\ldots, L$ and $f_y$ with $M$ components, weights $w_{y,1}, w_{y,2}, \ldots, w_{y,M}$, locations
\begin{equation}
    \vec{y}_j = \begin{bmatrix} y_{1,j}, y_{2,j}, \ldots, y_{D,j} \end{bmatrix}^T \in \NewR^D
\end{equation}
for $j=1,\ldots, M$, the Cram{\'e}r-von Mises distance $D$ is given by
\begin{equation}
    D = D_{yy} - 2 D_{xy} + D_{xx} + c \, D_{E} \enspace ,
    \label{Eq_CvD}
\end{equation}
with
\begin{equation}
    D_{yy} = \sum_{i=1}^M \sum_{j=1}^M w_{y,i} \cdot w_{y,j} \cdot \xlog \left( \sum_{d=1}^D \left( y_{d,i} - y_{d,j} \right)^2  \right)
    \enspace ,
\end{equation}
\begin{equation}
    D_{xy} = \sum_{i=1}^L \sum_{j=1}^M w_{x,i} \cdot w_{y,j} \cdot \xlog \left( \sum_{d=1}^D \left( x_{d,i} - y_{d,j} \right)^2 \right)
    \enspace ,
\end{equation}
\begin{equation}
    D_{xx} = \sum_{i=1}^L \sum_{j=1}^L w_{x,i} \cdot w_{x,j} \cdot \xlog \left( \sum_{d=1}^D \left( x_{d,i} - x_{d,j} \right)^2 \right)
    \enspace ,
\end{equation}
with $\xlog(z) = z \cdot \log(z)$. $D_{E}$ can be viewed as a penalty term (with weight $c$) that ensures equal means and is given by
\begin{equation}
    D_{E} = \sum_{d=1}^D \left( \sum_{i=1}^L w_{x,i} \cdot x_{d,i} - \sum_{i=1}^M w_{x,i} \cdot y_{d,i} \right)^2
    \enspace .
\end{equation}
%
%
When only the minimizer w.r.t.\ parameters of $f_x$ is desired, but not the corresponding value of $D$, $D_{yy}$ can be neglected.

\subsection{Map Optimization}

%
%
In this paper, $L$ and $M$ are assumed to be equal, with $w_{x,i}=w_{e,i}$, $w_{y,i}=\tilde{w}_{e,i}$, and $\vec{x}_i=\vec{x}_{e,i}$, $\vec{y}_i=\tilde{\vec{x}}_{e,i}$, $i=1,\ldots, L$.
This leads to the following dependency of $D$
\begin{equation}
    D = D\bigl( {\cal W}_e, {\cal X}_e, \tilde{{\cal W}}_e, \tilde{{\cal X}}_e \bigr) \enspace .
\end{equation}
As ${\cal X}_e$ is obtained from $\tilde{{\cal X}}_e$ via the map ${\cal X}_e = \vec{M}(\tilde{{\cal X}}_e)$, this can be rewritten as
\begin{equation}
    D = D\bigl( {\cal W}_e, \vec{M}(\tilde{{\cal X}}_e), \tilde{{\cal W}}_e, \tilde{{\cal X}}_e \bigr) \enspace .
\end{equation}
The optimal map $\vec{M}^\ast$ is now found by minimization
\begin{equation}
    \vec{M}^\ast = \argmin_{\vec{M} \in {\cal M}} D\bigl( {\cal W}_e, \vec{M}(\tilde{{\cal X}}_e), \tilde{{\cal W}}_e, \tilde{{\cal X}}_e \bigr) \enspace ,
\end{equation}
where ${\cal M}$ is the set of viable maps.

%
%
The gradient of $D$ in \Eq{Eq_CvD} is available in closed form \cite{hanebeckOptimalReductionMultivariate2015}.
When the likelihood is given in analytic form and differentiable, the gradient with respect to the map parameters can be derived.
A BFGS quasi-Newton method is used for optimization.

\section{Progressive Processing} \label{Sec_ProgressiveProcessing}

%
%
In the previous section, we assumed that reweighting of particles with the likelihood $f_L(\vec{x})$ kept all particles ``alive''.
%
%
Usually, however, performing the Bayes update in one step leads to particle degeneration.
Only a few samples stay ``alive'', the rest is close to zero.
In that case, resolution is lost as not all particles contribute to the density representation.

%
%
A proven remedy for keeping particles ``alive'' is to perform progressive processing \cite{ruoffProgressiveCorrectionDeterministic2011}.
%
%
The likelihood is decomposed into a product of sub-likelihoods, each of which is carefully selected to avoid degeneration
\begin{equation}
    f_L(\vec{x}) = f_L^{(1)}(\vec{x}) \cdot f_L^{(2)}(\vec{x}) \ldots f_L^{(K)}(\vec{x})
    = \prod_{k=1}^K f_L^{(k)}(\vec{x}) \enspace .
\end{equation}
%
%
%
As we use a product decomposition, each sub-likelihood is intuitively ``wider'' than the original one.
%
%
Sequential Bayes sub-updates with the sub-likelihoods then provide the desired posterior:

\medskip
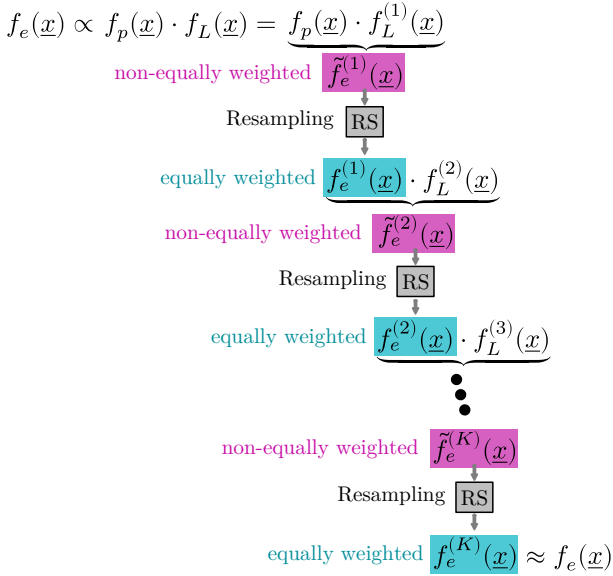
\includesvg{Graphics/BlockDiagrams_updateChain.svg}
\medskip

%
%
After every sub-update, we obtain an non-equally weighted sub-posterior $\tilde{f}_e^{(k)}(\vec{x})$.
In order to prepare for the next sub-update, the optimal resampling method from \Sec{Sec_Resampling} is used.
This involves mapping samples $\tilde{\vec{x}}_{e,i}^{(k)}$ to $\vec{x}_{e,i}^{(k)}$ with sub-mapping $\vec{M}^{(k)}(.)$.
This results in an equally weighted sub-posterior $f_e^{(k)}(\vec{x}) \approx \tilde{f}_e^{(k)}(\vec{x})$.
After $K$ sub-update steps, the result is an equally weighted sub-posterior $f_e^{(K)}(\vec{x})$, which is equal to $f_e(\vec{x})$.

%
%
The total map from prior samples $\vec{x}_{p,i}$ to posterior samples $\vec{x}_{e,i}$ is given by composition of the sequence of individual mappings as
\begin{equation}
    \vec{M}(\vec{x}) = \vec{M}^{(K)}\biggl( \vec{M}^{(K-1)}\Bigl( \ldots \vec{M}^{(2)}\bigl(\vec{M}^{(1)}(\vec{x}) \ldots\bigr) \Bigr) \biggr)
\end{equation}
or
\begin{equation}
    \vec{M} = \vec{M}^{(K)} \circ \vec{M}^{(K-1)} \circ \dots \circ \vec{M}^{(2)} \circ \vec{M}^{(1)} \enspace .
\end{equation}

\section{Numerical Results} \label{Sec_Numerical}

\subsection{Sanity Check: Linear System}

\begin{figure*}[h]
    \begin{center}
        \includesvg{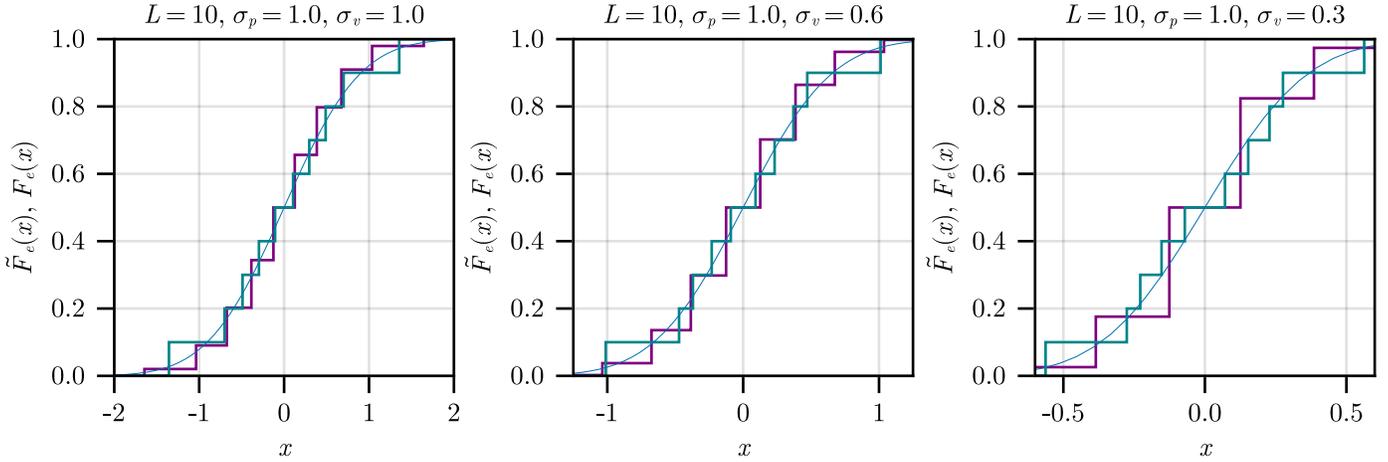}
    \end{center}
    \caption{%
        Results of a linear update with $L=10$ samples. The true continuous posterior is shown in blue.
        The non-equally weighted Dirac mixture $\tilde{f}_e(x)$ after multiplication of $f_p(x)$ with the likelihood $f_L(x)$ is shown in purple.
        The equally weighted Dirac mixture $f_e(x)$ produced by the proposed Bayesian particle flow is shown in turquoise.
        Please note the different x-axes scales.}
    \label{Fig_MapGeneration_L10}
\end{figure*}

\begin{figure*}[h]
    \begin{center}
        \includesvg{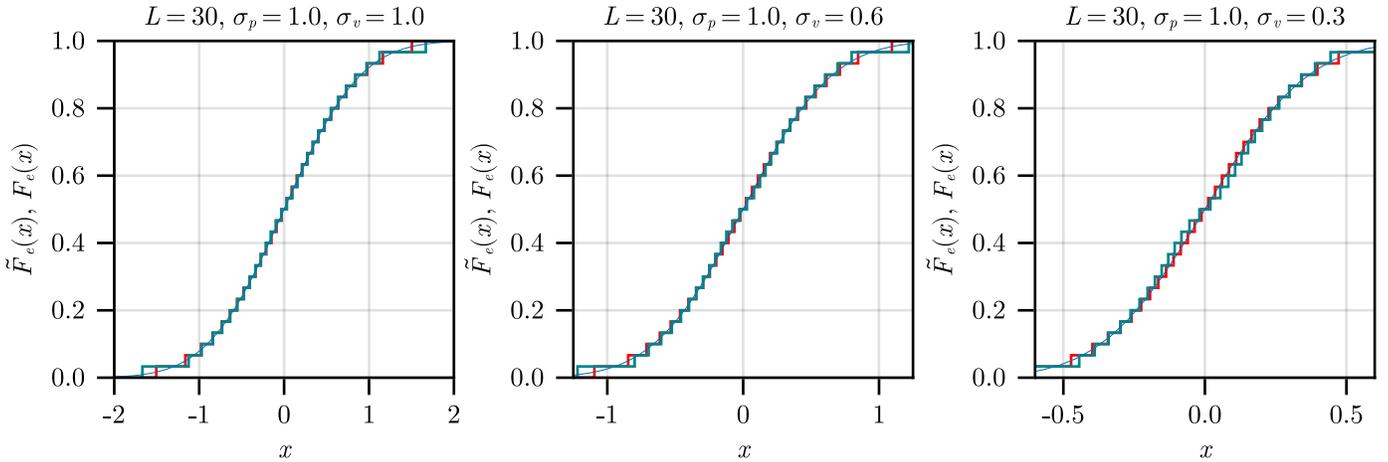}
    \end{center}
    \caption{%
        Results of a linear update with $L=30$ samples. The true continuous posterior is shown in blue.
        The equally weighted Dirac mixture $f_e(x)$ produced by the proposed Bayesian particle flow is shown in turquoise.
        The reference Dirac mixture obtained by directly sampling from the true continuous posterior is shown in red.
        Please note the different x-axes scales.}
    \label{Fig_MapGeneration_L30}
\end{figure*}

%
%
As a first example, we consider the simplest scalar linear measurement equation $y=x+v$.
$x$ is the desired state, $y$ is the measurement, and $v$ is Gaussian measurement noise with $v \sim f_v(v) = \N(v; 0, 1)$%
\footnote{$\N\left(x; m, \sigma \right)$ is a Gaussian density over realizations $x$ with mean $m$ and standard deviation $\sigma$.}.
The prior is Gaussian and given by $f_p(x)=\N(x; 0,1)$.
For $y=\hat{y}$, the likelihood is given by $f_L(x)=f(\hat{y} | x) = f_v(\hat{y}-x) = \N(\hat{y}; x, 1)$.

%
%
We assume that only samples $x_{p,i}$ of the prior density $f_p(x)$ are available.
The proposed method is used to perform the Bayesian filter step with the likelihood $f_L(x)$.
This produces samples $x_{e,i}$ of the posterior $f_e(x)$.

%
%
In this simple case, we can use the analytic prior and the likelihood to calculate the true posterior and its CDF.
The true posterior is given by $f_e^t(x) = \N(x; \hat{y}/2, 1/2)$.
Samples $x_{e,i}^t$ from $f_e^t(x)$ can now be used as reference for the samples $x_{e,i}$.
Note that the analytic prior and the reference samples are not known to the estimator.

%
%
For $L=10$ particles, prior standard deviation $\sigma_p=1$, and noise standard deviations $\sigma_v=1$, $\sigma_v=0.6$, and $\sigma_v=0.3$, we obtain the results in \Fig{Fig_MapGeneration_L10}.
The true continuous posterior (in blue) is closely approximated by the samples generated by the proposed Bayesian particle flow (in turquoise).
For smaller noise variances, the approximation gets worse.
This is expected due to the finite resolution of the prior density with only $L=10$ particles.

In \Fig{Fig_MapGeneration_L30}, the results of the same setup are shown for $L=30$ particles.
The true  continuous posterior is again shown in blue.
It is almost perfectly approximated by the samples generated by the proposed Bayesian particle flow (in turquoise).
For smaller noise variances, the approximation gets only slightly worse.
As a comparison, the reference samples $x_{e,i}^t$ obtained by directly sampling from the true continuous posterior are shown in red.

%
%
The corresponding total map from prior $\rv{x}_p$ to posterior $\rv{x}_e$ is shown in \Fig{Fig_linearMap} (in red).
It is compared with the true linear map (in green).
The maps are almost identical in the relevant region.

\subsection{Cubic Sensor Problem}

%
%
We now consider the (in-)famous cubic sensor problem with measurement equation $y=x^3+v$.
Again, $x$ is the state, $y$ the measurement, and $v$ Gaussian measurement noise with $v \sim f_v(v) = \N(v; 0,\sigma_v)$.
The prior is Gaussian and given by $f_p(x)=\N(x; 0, 1)$.
For $y=\hat{y}$, the likelihood is given by $f_L(x)=f_v(\hat{y}-x^3) = \N\left(\hat{y}; x^3, \sigma_v\right)$.
The true posterior is given by
\begin{equation}
    \tilde{f}_e(x) \propto f_p(x) \cdot f_L(x) = \N(0,1) \cdot \N\left(\hat{y}; x^3,\sigma_v\right) \enspace .
\end{equation}

\begin{YellowBox}
    \begin{remark}
        Sampling from $\tilde{f}_e(x)$ is difficult.
        Calculating the required CDF and its inverse can only be done by numerical integration.
    \end{remark}
\end{YellowBox}

\begin{figure}
    \begin{center}
        \includesvg{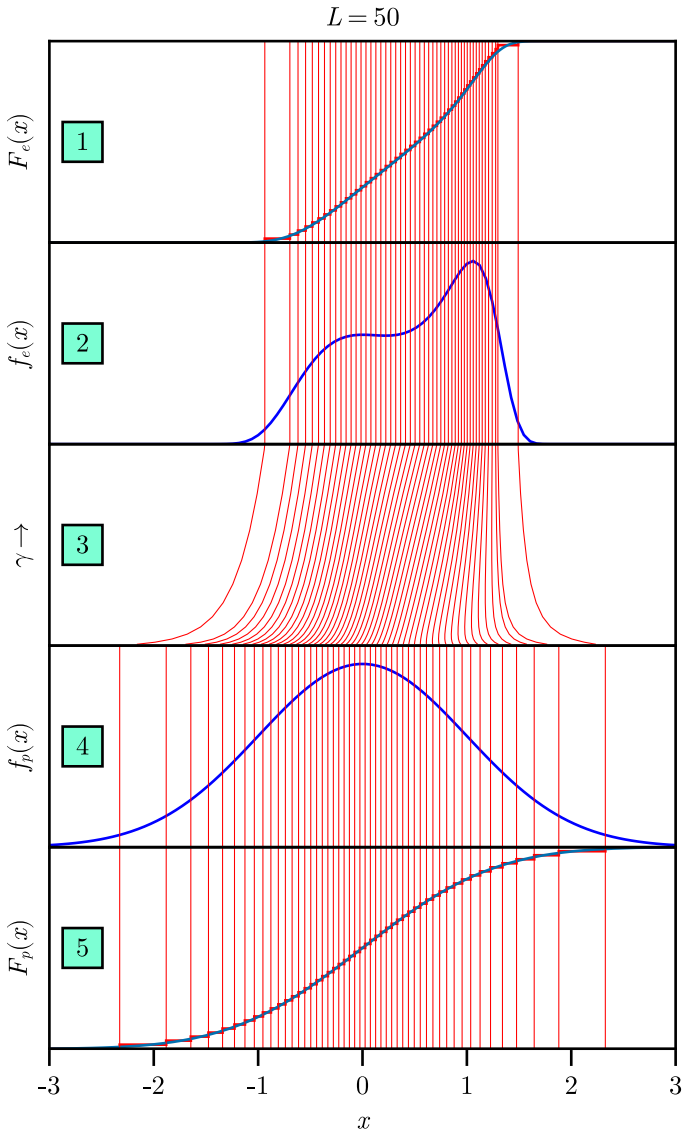}
    \end{center}
    \caption{%
        Cubic sensor problem: Results of update.
        (1,2) Posterior $f_e(x)$, cumulative $F_e(x)$, and samples $x_{e,i}$.
        (3) Flow of particles.
        (4,5) Prior $f_p(x)$, cumulative $F_p(x)$, and samples $x_{p,i}$.
    }
    \label{Fig_cubicSensor_20_full}
\end{figure}

%
%
We are given only samples $x_{p,i}$ of the prior density $f_p(x)$ and the analytic likelihood $f_L(x)$.
Samples $x_{e,i}$ of the posterior $f_e(x)$ are calculated with the proposed Bayesian flow.
%
%
\Fig{Fig_cubicSensor_20_full}~(4, 5) show the prior $f_p(x)$, the cumulative $F_p(x)$, and its samples $x_{p,i}$.
\Fig{Fig_cubicSensor_20_full}~(3) shows the flow of the particles.
\Fig{Fig_cubicSensor_20_full}~(1, 2) show the posterior $f_e(x)$, the cumulative distribution $F_e(x)$, and its samples $x_{e,i}$.
%
%
The total mapping from $\rv{x}_p$ to $\rv{x}_e$ is shown in \Fig{Fig_cubicSensorMap} (in red) compared to numerical reference (in green).
Again, the maps are almost identical in the relevant region.

\begin{figure}
    \begin{center}
        \includesvg{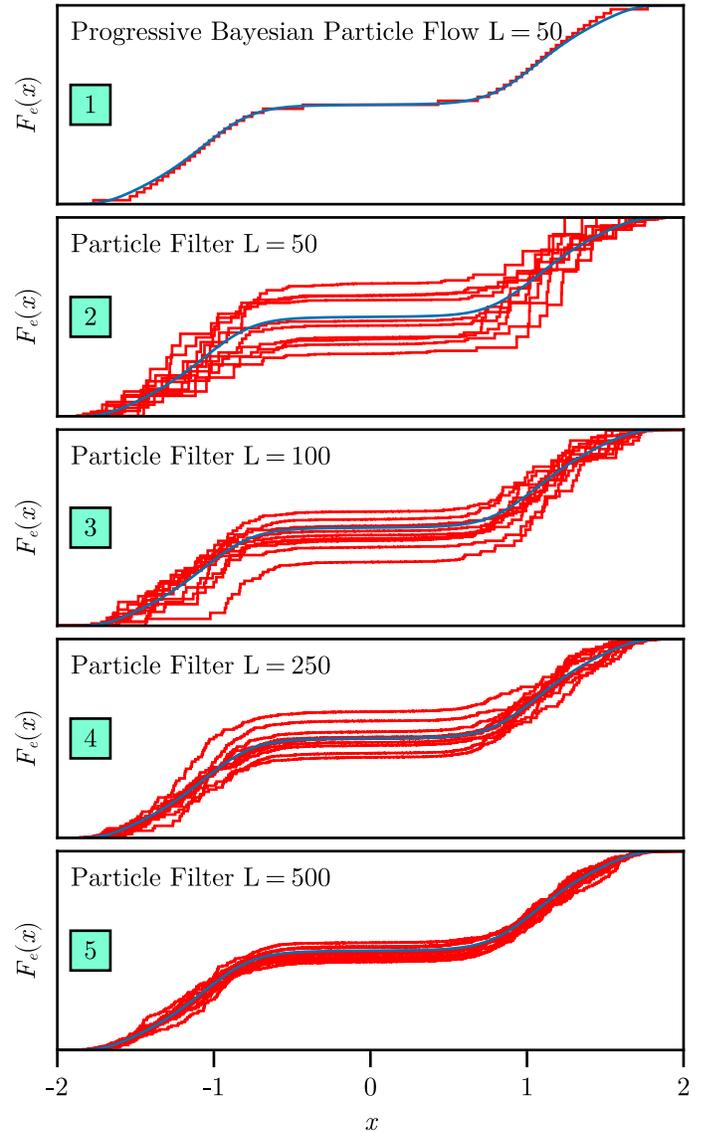}
    \end{center}
    \caption{Comparing the Bayesian particle flow with standard particle filter for single measurement update with likelihood shown in \Fig{Fig_strangeDensity}. (1) Result of proposed filter for $L=50$. (2)-(5) Results of particle filter for different $L$ with $10$ runs each.}
    \label{Fig_compareParticleFilter}
\end{figure}

\subsection{Comparison with Particle Filter}

\begin{figure*}
    \begin{minipage}{\textwidth}
        \begin{minipage}{.325\textwidth}
            \begin{center}
                \includesvg{Figures/linearUpdateMapping_L10_sv1.svg}
                \caption{Map $M(.)$ for linear filter step.}
                \label{Fig_linearMap}
            \end{center}
        \end{minipage}%
        \hspace*{\fill}
        \begin{minipage}{.325\textwidth}
            \begin{center}
                \includesvg{Figures/cubicSensorMapping_L50.svg}
                \caption{Map $M(.)$ for cubic sensor problem.}
                \label{Fig_cubicSensorMap}
            \end{center}
        \end{minipage}
        \hspace*{\fill}
        \begin{minipage}{.325\textwidth}
            \begin{center}
                \includesvg{Figures/strangeDensity.svg}
                \caption{Density in \Eq{Eq_strangeLikelihood} used for comparison with the standard particle filter.}
                \label{Fig_strangeDensity}
            \end{center}
        \end{minipage}
    \end{minipage}
\end{figure*}

%
%
The proposed Bayesian particle flow will now be compared with the standard particle filter.
The underlying continuous prior is given by $f_p(x)=\N(x; 0, 1)$.
The likelihood
    {\small
        \begin{equation}
            f_L(x) = \exp\left( \! \!-\frac{1}{2} \bigl( (x-1.2)(x-1.5)(x+1.2)(x+1.5) \bigr)^2 \! \right)
            \label{Eq_strangeLikelihood}
        \end{equation}}%
is shown in \Fig{Fig_strangeDensity} with the prior $f_p(x)$ shown in blue, the likelihood $f_L(x)$ in green, and the posterior $f_e(x)$ in red.
These underlying continuous densities are unknown to the estimator.

%
%
For the simulation, we only have samples $x_{p,i}$, $i=1,\ldots, L$ of the prior $f_p(x)$ and the analytic likelihood in \Eq{Eq_strangeLikelihood} available.
%
%
The results are shown in \Fig{Fig_compareParticleFilter}.
The posterior estimate of the proposed Bayesian particle flow for $L=50$ particles in \Fig{Fig_compareParticleFilter}~(1) is very close to the true posterior.
Results of the standard particle filter are shown in \Fig{Fig_compareParticleFilter}~(2)-(5) for different $L$ and ten runs each.
For $L=50$, the results are of low resolution at the peaks of the posterior.
This is due to the fact that the filter step  of the particle filter produces weighted samples.
In addition, there is large variability between different runs.
For increasing $L$, the resolution degradation becomes less pronounced and the variability between runs decreases.
However, even for $L=500$ the variability is still clearly present.

\section{Conclusions} \label{Sec_Conclude}

%
%
A new Bayesian particle flow has been derived that deterministically guides particles from prior to posterior.
It does not require any continuous density representations, neither in its derivation nor in its implementation.
%
%
It is composed of a finite sequence of potentially non-monotonic maps for propagating particles.
%
%
The filter works with arbitrary nonlinear measurement equations.
However, it is assumed that it has already been converted to a likelihood function that can be evaluated.
%
%
The method is easy to understand and its implementation is straightforward.

%
%
Several distance measures could be employed for map optimization, see \SubSec{SubSec_PropertiesMap}.
However, we found the distance derived in \cite{hanebeckOptimalReductionMultivariate2015} most useful as it has low complexity and is differentiable.
Its complexity is quadratic in the number of particles $L$ and linear in the number of dimensions $D$.

%
%
Several aspects have been omitted in this paper due to space restrictions.
%
%
%
(i)~We did not give details on how to find appropriate sub-likelihoods that keep particles ``alive'' and their number.
For that purpose, we use the method described in \cite{steinbringProgressiveGaussianFiltering2014}.
%
%
(ii)~We assumed that the number of particles $L$ is constant.
This is not necessary.
The number of samples can be adapted to the complexity of the underlying density.
Methods for an efficient adaptation will be developed.
%
%
(iii)~We did not discuss the prediction step, i.e., propagating particles through the system model.
For deterministic particles, the prediction step is significantly different from the random case in terms of combining state and noise samples.
In order to avoid a full Cartesian product, the method from \cite{eberhardtOptimalDiracApproximation2010} can be used.


\printbibliography

\end{document}